\documentclass{article}

\usepackage[final]{corl_2026} %

\usepackage{amssymb,amsmath}
\usepackage{cite} %
\usepackage{booktabs}
\usepackage[font=footnotesize,labelformat=simple]{subcaption}
\usepackage[font=footnotesize]{caption}
\usepackage{needspace}
\usepackage{afterpage}

\newcommand{\methodname}{\textit{SGS}}

\newcommand{\taskname}[1]{\texttt{#1}}
\newcommand{\taskNut}{\taskname{nut-and-bolt assembly}}
\newcommand{\taskRod}{\taskname{rod-in-hole insertion}}
\newcommand{\taskPeg}{\taskname{rectangular-peg-in-hole insertion}}
\newcommand{\taskBNC}{\taskname{BNC assembly}}
\newcommand{\taskGear}{\taskname{gear mesh insertion}}
\newcommand{\taskWaterproof}{\taskname{waterproof insertion}}

\usepackage{thmtools,thm-restate}
\usepackage[noend, ruled, linesnumbered]{algorithm2e}
\usepackage[noend]{algpseudocode}
\usepackage{multirow}
\usepackage{tabulary}
\usepackage{graphicx}
\usepackage{tikz}
\usepackage{wrapfig}
\usepackage{float}  %
\usepackage{natbib}
\usepackage{multirow}
\usepackage{makecell}

\usepackage{xcolor}

\usepackage{hyperref}
\hypersetup{bookmarksopen,bookmarksnumbered,
pdfpagemode=UseOutlines,
colorlinks=true,
linkcolor=blue,
anchorcolor=blue,
citecolor=blue,
filecolor=blue,
menucolor=blue,
urlcolor=blue
}

\title{
A Balanced Data Diet: Addressing the Exploration Bottleneck in Mega-Scale RL for Robot Control}

\author{
Octi Zhang$^{1,2*}$ \quad
Mateo Guaman Castro$^{1,*}$ \quad
Patrick Yin$^{1,*}$ \\[4pt]
\bf Ignacio Dagnino$^{1}$ \quad
Abhishek Gupta$^{1}$ \quad
Rosario Scalise$^{1,*,\dagger}$ \quad
Byron Boots$^{1, \dagger}$ \\[4pt]
\normalfont $^{1}$University of Washington \quad
$^{2}$NVIDIA \quad
$^{*}$Equal contribution \quad
$^{\dagger}$Equal advising
}

\begin{document}
\maketitle
\vspace{-24pt}
\begin{figure}[H]
  \centering
  \makebox[\linewidth][c]{\includegraphics[width=1\linewidth]{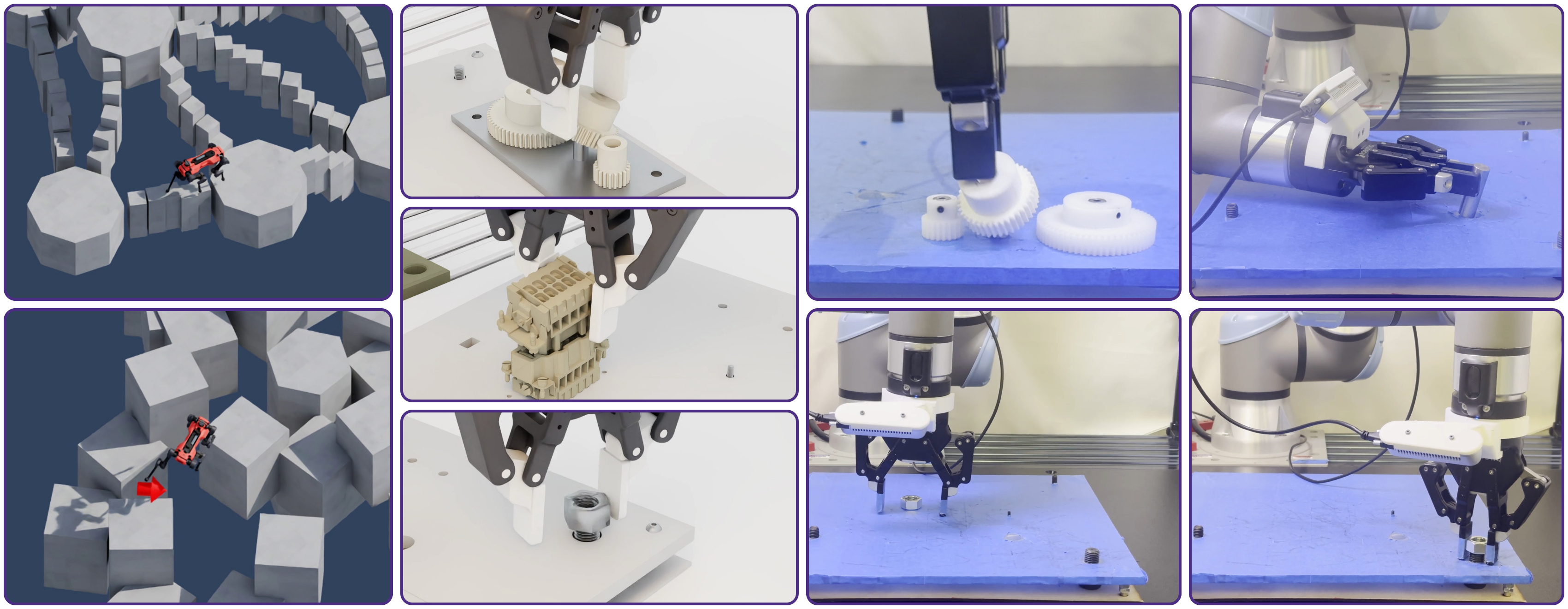}}
  \caption{\textbf{\methodname{} enables large-scale RL to learn challenging robotic behaviors, from traversing custom-designed difficult terrains to dexterous, contact-rich assembly.} By focusing training on configurations at the frontier of the policy's capabilities, \methodname{} solves tasks that prior sampling methods struggle to learn. We further distill the manipulation policies into RGB-based policies and transfer them zero-shot to real hardware.}
  \label{fig:1_hero}
\end{figure}

\begin{abstract}
General-purpose robots must perform a wide range of tasks from agile locomotion to dexterous manipulation. While sim-to-real reinforcement learning (RL) has proven to be a useful tool for this goal, current RL pipelines depend on engineering-heavy, per-task structural priors such as shaped rewards and demonstrations.  Recent work has shown that diverse simulator resets, combined with massively parallel simulation, can alleviate much of this engineering burden on several manipulation problems. However, we find that naively scaling this paradigm to more precise or dynamic problems remains non-trivial. While simulator resets can help with exploration, uniformly sampling over this distribution wastes a growing fraction of learning experience on task configurations the policy has already mastered or cannot yet attempt. This makes it challenging to see the expected benefits of scaling parallel environments for RL, since much of the learning signal in a batch is wasted during learning. To mitigate this, we introduce Success Guided Sampling (\methodname{}), a simple adaptive sampler that concentrates RL training on task configurations around the frontier of the policy's capabilities. Doing so allows large-scale simulated RL to make the most out of the experience in a batch, enabling much more effective scaling to large-scale parallel simulation. Across experiments using up to $2^{20}$ (over one million) parallel environments, \methodname{} enables RL to solve challenging multi-terrain quadruped locomotion and contact-rich assembly tasks that prior methods fail to solve. Finally, we distill the learned manipulation policies into RGB-based policies and demonstrate zero-shot transfer to several challenging assembly tasks on real hardware. Project website: \url{https://sgs-rl.github.io/}.

\end{abstract}


\clearpage

\section{Introduction}
\label{sec:introduction}

General-purpose robotics requires a training recipe that works across diverse embodiments and tasks, from legged systems navigating complex terrain to dexterous arms manipulating contact-rich objects. Sim-to-real reinforcement learning with massively parallel physics simulation~\citep{todorov2012mujoco, mittal2023orbit, makoviychuk2021isaacgym, gu2023maniskill2} has emerged as a promising pathway toward this goal, producing policies that transfer to real hardware across agile locomotion~\citep{rudin2022learning, hwangbo2019learning, cheng2024extreme}, dexterous manipulation~\citep{handa2023dextreme, singh2024dextrah, akkaya2019solving}, and contact-rich assembly~\citep{narang2022factory, yin2026emergent}. 

However behind these successes of RL lies substantial per-task engineering. Reward functions require expert tuning and frequently induce unintended behaviors ~\citep{he25viral, rudin2025parkour}. Curricula, whether hand-designed ~\citep{florensa2017reverse} or automatic within a fixed parameterization ~\citep{graves2017automated, florensa2018automatic, handa2023dextreme}, are tuned or specified per task. Demonstrations, when used to bootstrap exploration, both constrain the resulting behavior and require collection effort ~\citep{peng2018deepmimic, rajeswaran2018learning, tao2024reverse, bauza25demostart}. And achieving breadth across tasks typically requires training single-task experts and distilling into a generalist ~\citep{rudin2025parkour}, bloating the training pipeline. Moreover, each of these choices induces a new set of hyperparameters that requires tuning. Each new task is a fresh engineering project, making this paradigm far from a turnkey solution. 

A growing body of work uses simulator resets to make difficult robotic tasks learnable, exposing policies to useful intermediate states rather than requiring successful exploration from scratch~\citep{florensa2017reverse, tao2024reverse, toussaint2026constrained, yin2026emergent}. These approaches construct starting states through demonstrations, geometric constraints, or programmatic sampling. OmniReset~\citep{yin2026emergent} shows that diverse, programmatically generated resets combined with massively parallel RL can substantially reduce task-specific reward and demonstration engineering. Together, these results establish reset distributions as an important component of scaling robot learning.

Providing useful reset states is only part of the problem: training must also allocate experience across them. Under uniform sampling, many environments may start from configurations the policy has already mastered or cannot yet solve, leaving fewer opportunities to practice behaviors it is still learning. In our experiments, this limits the gains from increasing the number of parallel environments, particularly on challenging manipulation and locomotion tasks.

We introduce \methodname{}, a simple sampler that adapts task-configuration sampling throughout training using non-parametric estimates of the policy's current success rates. Intuitively, learning signal is concentrated in regions of the space that are neither too hard nor too easy under the current policy. We instantiate this by weighting success estimates with a beta distribution, which prioritizes configurations with moderate success rates. This ensures that sampling is generically guided by the policy's performance rather than a manually ordered task curriculum. We study the same instantiation of \methodname{} across difficult-terrain quadruped whole-body locomotion and dexterous NIST assembly tasks~\citep{kimble2020benchmarking}, including threaded assembly, insertion, connector assembly, and gear meshing. We train without demonstrations, using one simple reward function shared across all locomotion tasks and another shared across all manipulation tasks, at scales of up to approximately one million parallel environments. \methodname{} enables improved learning throughput, solving much more challenging tasks at scale than possible before. 

We outline our contributions as follows:
\begingroup
\setlength{\itemsep}{0pt}
\setlength{\parsep}{0pt}
\setlength{\parskip}{0pt}
\setlength{\topsep}{0pt}
\setlength{\partopsep}{0pt}
\begin{enumerate}
    \item We introduce \methodname{}, a simple success-based sampler for large-scale robot RL, and study how adaptive sampling of task configurations scales to approximately one million parallel environments.
    \item We evaluate \methodname{} across difficult-terrain quadruped locomotion and diverse assembly tasks (Figure~\ref{fig:six_nist_tasks}). \methodname{} outperforms uniform sampling and Prioritized Level Replay (PLR)~\citep{jiang2021prioritized} on challenging multi-terrain whole-body locomotion and \taskNut{}, while performing comparably to uniform sampling on simpler tasks.
    \item We distill learned manipulation policies into RGB-based policies and demonstrate zero-shot transfer to several precise assembly tasks on real robot hardware (Figure~\ref{fig:real_transfer}).
\end{enumerate}
\endgroup

\section{Related Work}
\label{sec:related_work}
\textbf{Sim-to-real reinforcement learning.} Sim-to-real RL has enabled substantial advances in legged locomotion~\citep{rudin2022learning, kumar2021rma, hoeller2024anymal, rudin2025parkour, zhang2024learning, peng2018deepmimic, liao2025beyondmimic, li2025bfm} and dexterous manipulation~\citep{handa2023dextreme, akkaya2019solving, singh2024dextrah, narang2022factory, yin2026emergent}, often using PPO~\citep{schulman2017proximal}. These successes commonly rely on task-specific reward shaping, curricula, or demonstrations~\citep{rudin2025parkour, he2025viral, singh2024dextrah, akkaya2019solving, narang2022factory, tang2024automate}. We study a training recipe that combines diverse simulator resets, success-based adaptive sampling, and large-scale RL, using no demonstrations for RL training and one shared reward function within each domain.

\textbf{Reset distributions for exploration.} The choice of training-state distribution has long been recognized as important for policy learning~\citep{bagnell2003policy}. Resetting to useful intermediate states can make difficult exploration problems tractable, as demonstrated by backward learning from a single demonstration and Go-Explore~\citep{salimans2018montezuma, ecoffet2021first}. In robotics, reverse curricula, RFCL, and DemoStart use goal states or demonstration states to construct starting-state curricula~\citep{florensa2017reverse, tao2024reverse, bauza25demostart}, while sampling-based exploration and constrained sampling construct useful reset states from the geometry of manipulation~\citep{khandate2023sampling, toussaint2026constrained}. OmniReset generates diverse resets programmatically and combines them with massively parallel RL~\citep{yin2026emergent}. Building on this body of work, \methodname{} maintains empirical success estimates over a fixed set of task configurations and continuously adjusts how frequently each configuration is sampled during training.

\textbf{Scaling on-policy RL with parallel simulation.} Massively parallel simulation has made large-scale on-policy RL practical for robot learning~\citep{rudin2022learning, tao2024maniskill3}. Recent work improves learning at scale through parallel exploration and population-based training~\citep{singla2024sapg, petrenko2023dexpbt}, changes to loss estimation~\citep{beukman2026preventing}, and staggered resets that increase temporal diversity within training batches~\citep{bharthulwar2025staggered}. \methodname{} studies a complementary component of this training process: allocating parallel environments across task configurations according to the policy's current success rates.

\textbf{Automatic curricula and task sampling.} Automatic curriculum learning selects training tasks according to the learner's capabilities~\citep{narvekar2020curriculum}. Approaches include learning-progress-based task selection~\citep{graves2017automated, matiisen2019teacher}, asymmetric self-play that generates increasingly challenging goals~\citep{sukhbaatar2017intrinsic}, and unsupervised environment design. Within the latter, PAIRED generates environments using a regret objective, PLR prioritizes previously encountered levels based on TD-errors, and ACCEL evolves levels to construct a curriculum~\citep{dennis2020emergent, jiang2021prioritized, parker2022evolving}. Most closely related to \methodname{}, Sampling for Learnability (SFL) prioritizes levels using $p(1-p)$, where $p$ is empirical success rate~\citep{rutherford2024no}. \methodname{} follows the same principle of emphasizing moderate success rates, using rolling success estimates and a smooth beta-shaped weighting over a fixed configuration set. Its target success rate is adjustable, and every configuration retains nonzero sampling probability. We evaluate this simple approach against uniform sampling and PLR on our scaling benchmarks, with an additional SFL comparison on locomotion.

We do not claim a fundamentally new curriculum principle. Instead, we show that a simple variant of success-based sampling, combined with diverse resets and large-scale RL, can solve challenging locomotion and assembly tasks that prior sampling methods struggle to learn.

\section{Method}
\label{sec:method}
\subsection{Problem Setup}
\label{sec:problem_setup}

\textbf{Reinforcement Learning Problem:} We study the problem of learning goal-conditioned RL policies from scratch for challenging robotic control tasks.
We instantiate this problem as a goal-conditioned Markov Decision Process (MDP) $\mathcal{M} = (\mathcal{S}, \mathcal{A}, \mathcal{G}, P, r, \gamma, \rho)$ where $s \in \mathcal{S}$ is the state, $a \in \mathcal{A}$ is the robot's action, $g \in \mathcal{G}$ is the robot's goal, $P$ is the transition dynamics, $r$ is the reward function, $\gamma$ is the discount factor, and $\rho$ is the distribution over initial state and goal pairs $(s_0, g) \in \mathcal{S} \times \mathcal{G}$.
$s_0$ captures all episode-level environment variation, including terrain geometry for locomotion and object instances for manipulation.
The reward $r(s_t, a_t, g)$ is shared within each domain: one reward covers all locomotion tasks and another covers all manipulation tasks (\autoref{eq:reward_form}).
The objective is to maximize the expected discounted sum of rewards $J(\pi) = \mathbb{E}_{(s_0, g) \sim \rho, a \sim \pi}\left[ \sum_{t=0}^\infty \gamma^t r(s_t, a_t, g)\right]$.
We focus our RL training on compact Lagrangian states.

\textbf{Task configurations.} We represent each task configuration as $\tau=(s_0,g,e)$, where $s_0$ is the initial state, $g$ is the goal, and $e$ specifies the environment configuration. Each episode starts from one such configuration, and \methodname{} adapts how frequently configurations are sampled as the policy learns.

For locomotion, a task configuration specifies the terrain type and geometry, the robot's starting pose, and its goal pose. Sampling configurations therefore determines both which terrain the robot practices on and where it must travel within that terrain.

For manipulation, a task configuration specifies the assembly task, the initial robot and object states, and the receptacle pose. The goal is the assembled object pose relative to the receptacle, whose pose is represented in the policy's observations. Initial states cover Reaching, Stable Grasp, and Near-Goal resets, following OmniReset~\citep{yin2026emergent}. Sampling configurations determines which starting conditions the policy practices to complete the assembly.

In both domains, \methodname{} allocates training episodes across these configurations using the policy's current success rates. Appendices~\ref{appendix:locomotion_task_config} and~\ref{appendix:manipulation_task_config} provide the task-specific configuration and reset definitions.

\subsection{\methodname: Success Guided Sampling}

\begin{wrapfigure}{r}{0.45\textwidth}
    \centering
    \includegraphics[width=\linewidth]{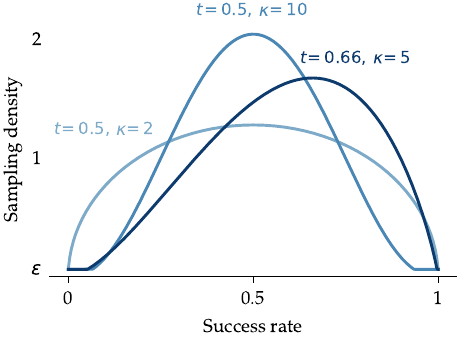}
    \caption{\textbf{Beta-shaped sampling distribution.} Different parameterizations prioritize coverage and sampling greediness.}
    \label{fig:beta_dist}
    \vspace{-1em}
\end{wrapfigure}

The core insight of \methodname~is that learning complex control requires exposing the policy to the right experience at the right time. In a goal-conditioned setting, success rate is a cheap, online signal of which configurations the policy should practice now. The aim is to spend most environment steps on configurations that are neither too easy nor too hard, while occasionally revisiting all of them to keep success-rate estimates accurate.

Before training, we sample a fixed set of $N$ configurations $\{\tau_i\}_{i=1}^{N}$ from the task distribution. At each episode reset, \methodname{} selects a configuration from this set. For each configuration, we maintain the last $H$ Boolean outcomes and estimate its success rate $p_i$ as their mean.

A configuration's success rate reflects both its difficulty and the policy's training history. It can change as the policy practices that configuration or learns transferable skills from others. \methodname{} therefore updates its estimates throughout training, concentrating experience on configurations with moderate success rates while continuing to revisit easier and harder configurations.

We then score each configuration with a Beta-shaped kernel in mode-concentration form, $\mathrm{Beta}(1+\kappa t,\; 1+\kappa(1-t))$, parameterized by a target success rate (i.e.\ mode) $t \in [0, 1]$ and concentration $\kappa \geq 0$:
\begin{equation}
    w_i \;=\; (p_i + \epsilon)^{\kappa t}\,(1 - p_i + \epsilon)^{\kappa(1-t)},
    \qquad
    \ell_i \;=\; \log\!\bigl(w_i + \epsilon\bigr),
    \label{eq:gps_logit}
\end{equation}
where $w_i$ is the unnormalized density at $p_i$ and $\ell_i$ the corresponding logit. The small $\epsilon > 0$ both prevents singularities at $p_i \in \{0,1\}$ and keeps every configuration's sampling probability non-zero, so that success-rate estimates can be updated even for configurations the robot cannot yet solve. At the end of each episode, the next configuration is sampled i.i.d.\ from a softmax over $\{\ell_i\}$ with temperature $T$:
\begin{figure}[t]
  \centering
  \includegraphics[width=\linewidth]{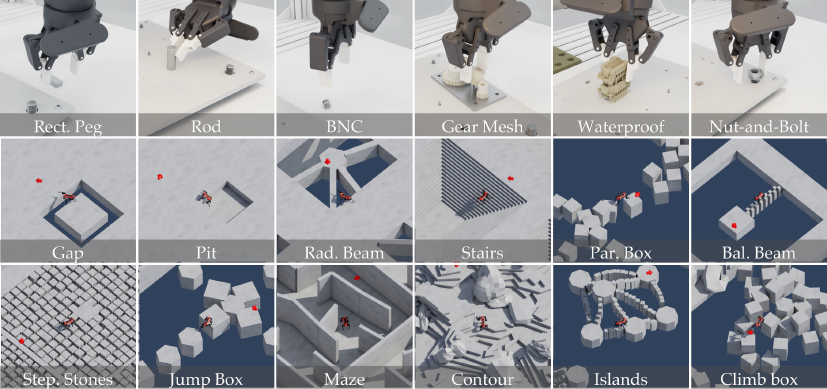}
  \caption{\textbf{Manipulation and locomotion tasks.} 6  NIST Assembly Task-Board (ATB1) \citep{kimble2020benchmarking} tasks and 12 of the 13 challenging locomotion terrains, solved with \methodname{}. Manipulation tasks require precision and dexterity to solve tasks with tight tolerances, such as screwing a nut onto a bolt. Locomotion tasks require the robot to string together multiple agile maneuvers as well as precise foothold placements to traverse complex terrains such as uneven floating islands.}
  \label{fig:six_nist_tasks}
  \label{fig:locomotion_tasks}
\end{figure}

\begin{equation}
    p(i) \;=\; \frac{\exp(\ell_i / T)}{\sum_{j=1}^{N} \exp(\ell_j / T)}
\end{equation}
Intuitively, configurations with very high or very low success rates are sampled with small but non-zero probability, while those near the target success rate are sampled most often (Figure~\ref{fig:beta_dist}). In practice, we use $N=32,768$ for manipulation and $N=104,000$ for locomotion, $H=100$, and $T=2$. We use $t=0.66$, $\epsilon=10^{-8}$, and $\kappa=5$ for locomotion, and $t=0.5$, $\epsilon=10^{-4}$, and $\kappa=1$ for manipulation. We find \methodname~to not be sensitive to most of these hyperparameters. Appendix~\ref{app:sgs_sensitivity} reports sensitivity to these choices and implementation details.

\subsection{Algorithmic Decisions for RL Training}
\label{sec:rl_design}

Beyond the sampler itself, \methodname~makes two deliberate algorithmic choices that work together to support training across a wide task configuration space with PPO~\citep{schulman2017proximal}.

\textbf{Shared rewards within each domain.} We use one simple reward function across all locomotion tasks and another across all manipulation tasks. Each combines a boolean task success reward obtained at the end of the episode with relatively small robot-specific, per-step regularization terms:
\begin{equation}
  r(s_t, a_t, g) \;=\;
  \lambda_{\text{task}}\, r_{\text{task}}(s_t, a_t, g)
  \;+\;
  \lambda_{\text{reg}}\, r_{\text{reg}}(s_t, a_t).
  \label{eq:reward_form}
\end{equation}
 The shared-reward design avoids specifying a different reward function for each terrain or assembly task. Appendix~\ref{appendix:reward_specification} gives the terms and weights for each domain.

\textbf{Scaling parallel environments.} When combined with adaptive sampling, we find that scaling the number of parallel environments enables \methodname~to solve substantially harder tasks than would otherwise be possible. Intuitively, when many task configurations produce no useful gradient signal (because the policy has already mastered them or cannot yet solve them), uniform sampling wastes a growing fraction of environments on these configurations as scale grows. \methodname~addresses this by concentrating rollouts on configurations where the policy is actively learning, so additional environments translate into denser learning signal rather than diminishing returns. This complements recent work on optimizer-side fixes to PPO saturation, including approaches addressing data redundancy from a single Gaussian policy~\citep{singla2024sapg} and, in recent work, degraded loss estimates at scale~\citep{beukman2026preventing}.

\section{Experimental Results}
\label{sec:result}

Through a series of large-scale scaling studies, we ask: \textbf{(Q1) Scaling}~--- does \methodname~scale better than uniform sampling and PLR~\citep{jiang2021prioritized} as parallel training environments grow? \textbf{(Q2) Capability}~--- does \methodname~unlock robotics tasks beyond what prior methods could solve, even at smaller scales? \textbf{(Q3) Transfer}~--- can dexterous manipulation behaviors learned with \methodname{} transfer zero-shot to real hardware?

\subsection{Locomotion}
\begin{figure}[t]
  \centering
  \includegraphics[width=0.2\linewidth,trim={420bp 60bp 268bp 100bp},clip]{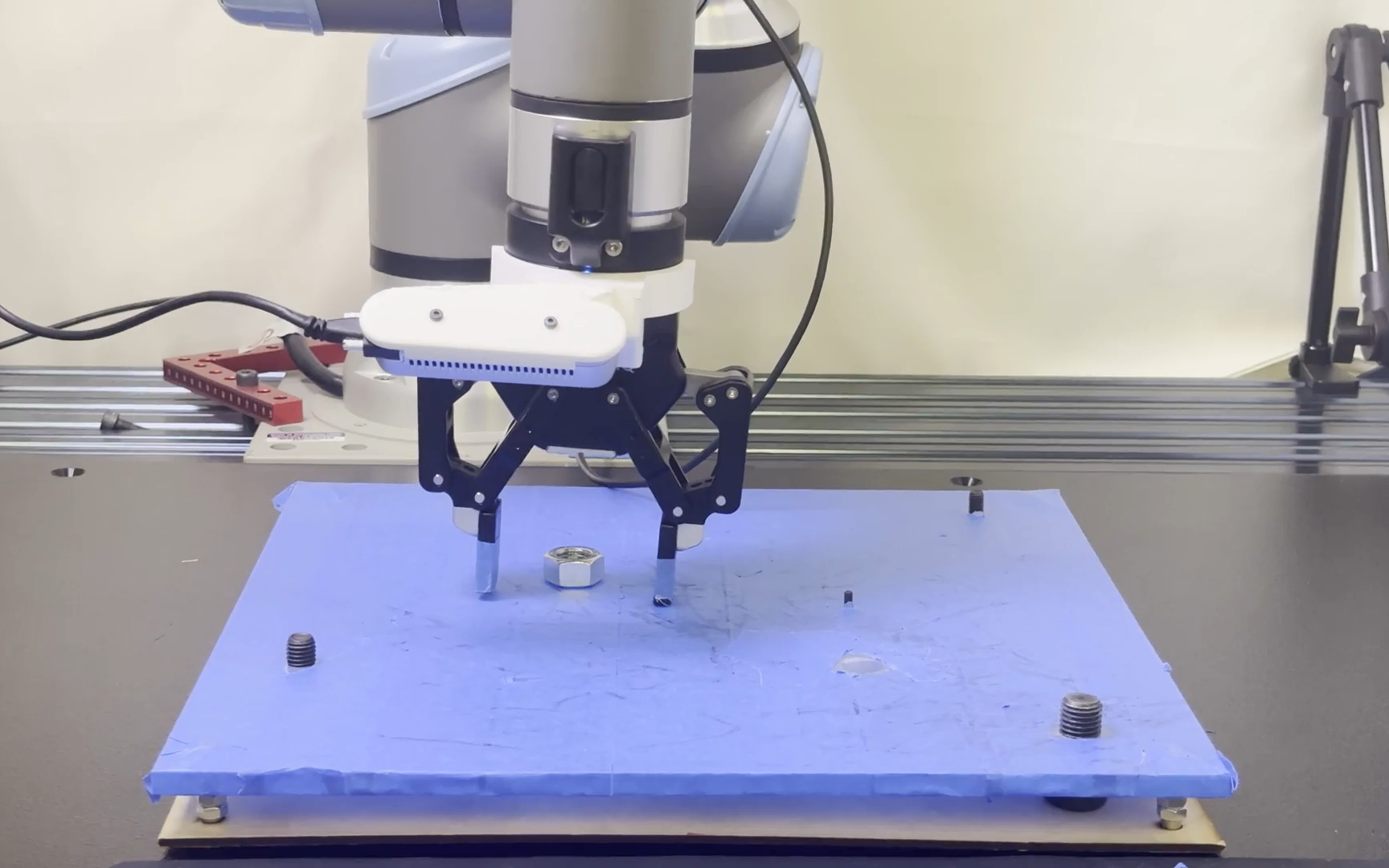}%
  \includegraphics[width=0.2\linewidth,trim={660bp 60bp 28bp 100bp},clip]{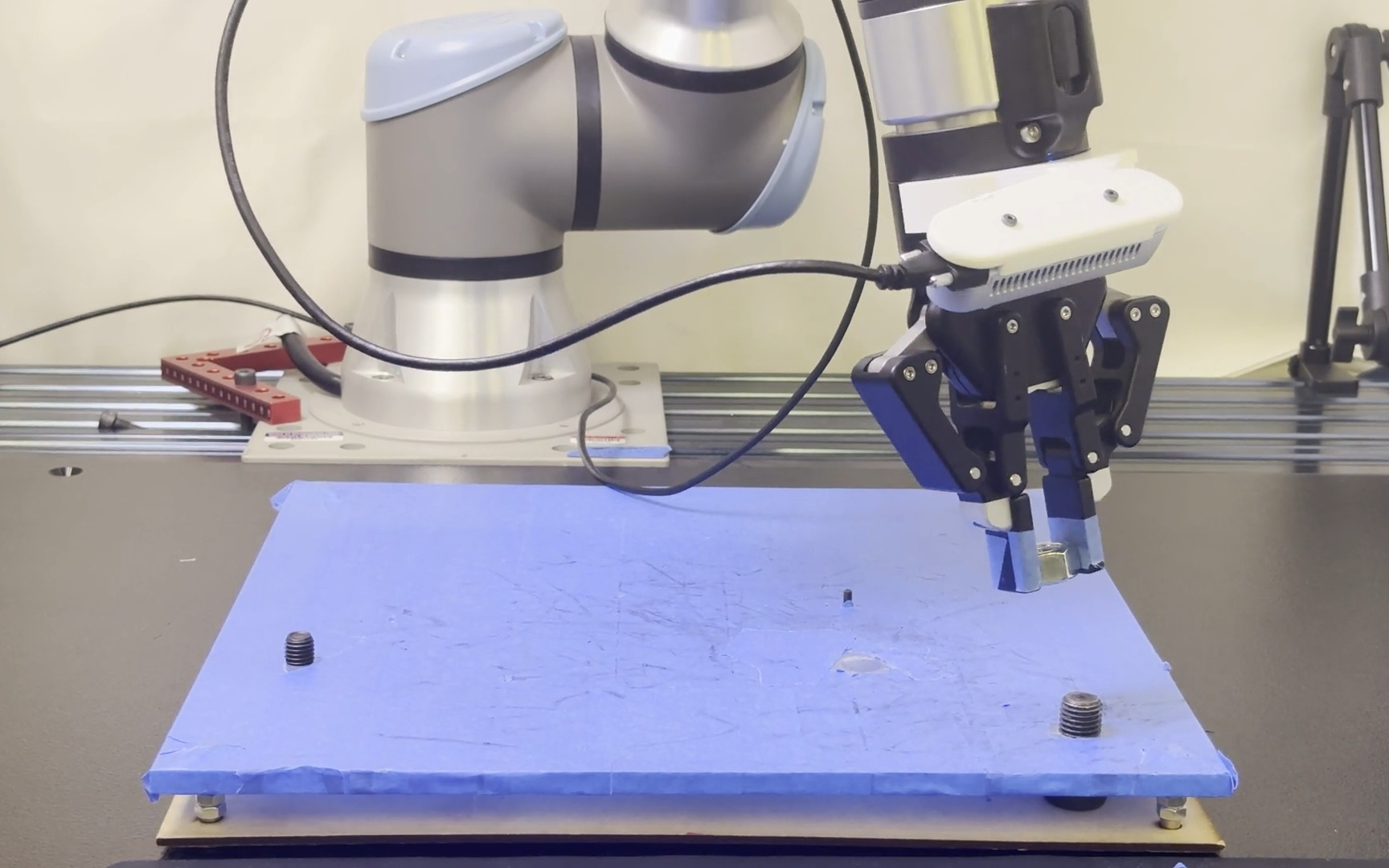}%
  \includegraphics[width=0.2\linewidth,trim={660bp 60bp 28bp 100bp},clip]{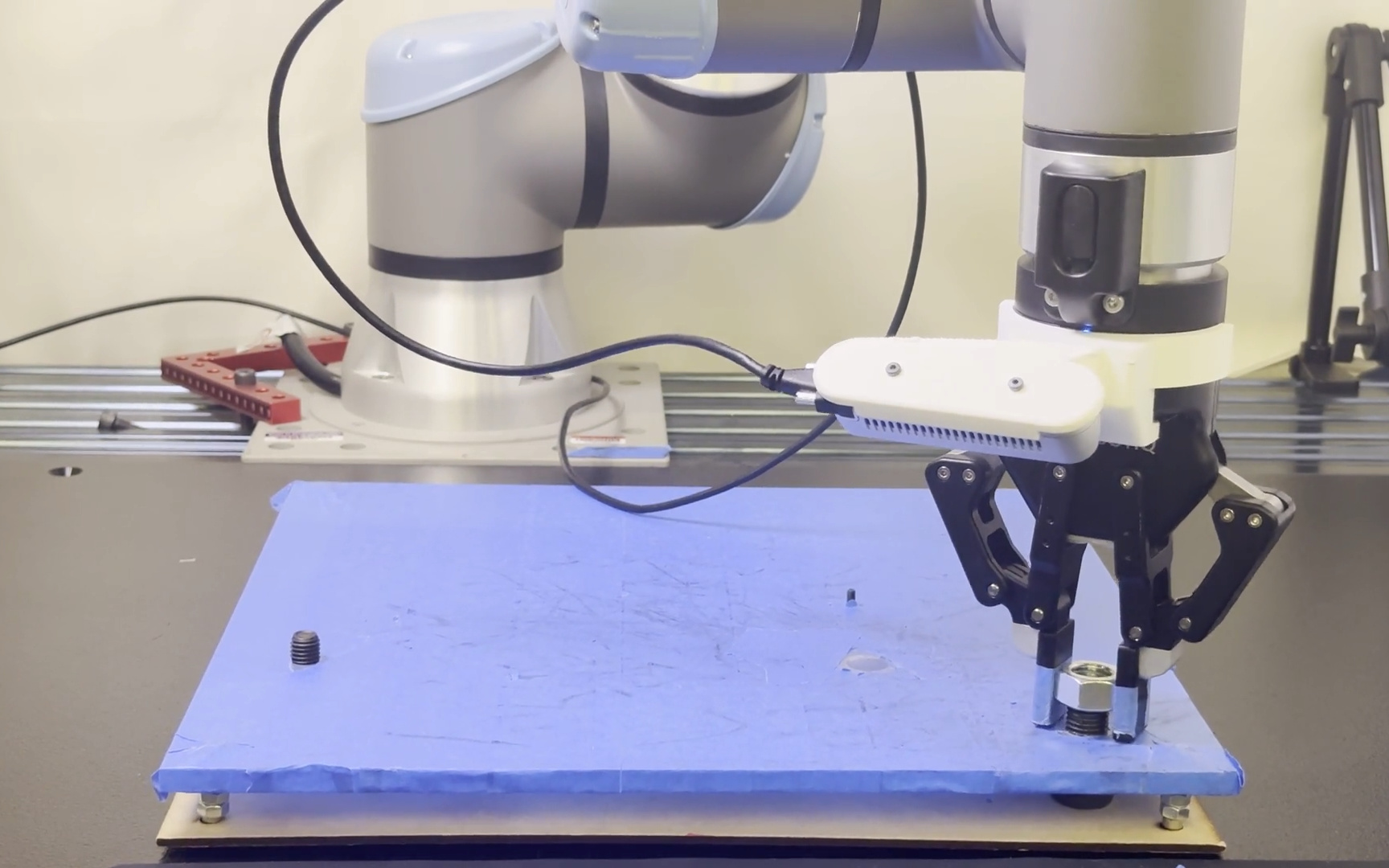}%
  \includegraphics[width=0.2\linewidth,trim={660bp 60bp 28bp 100bp},clip]{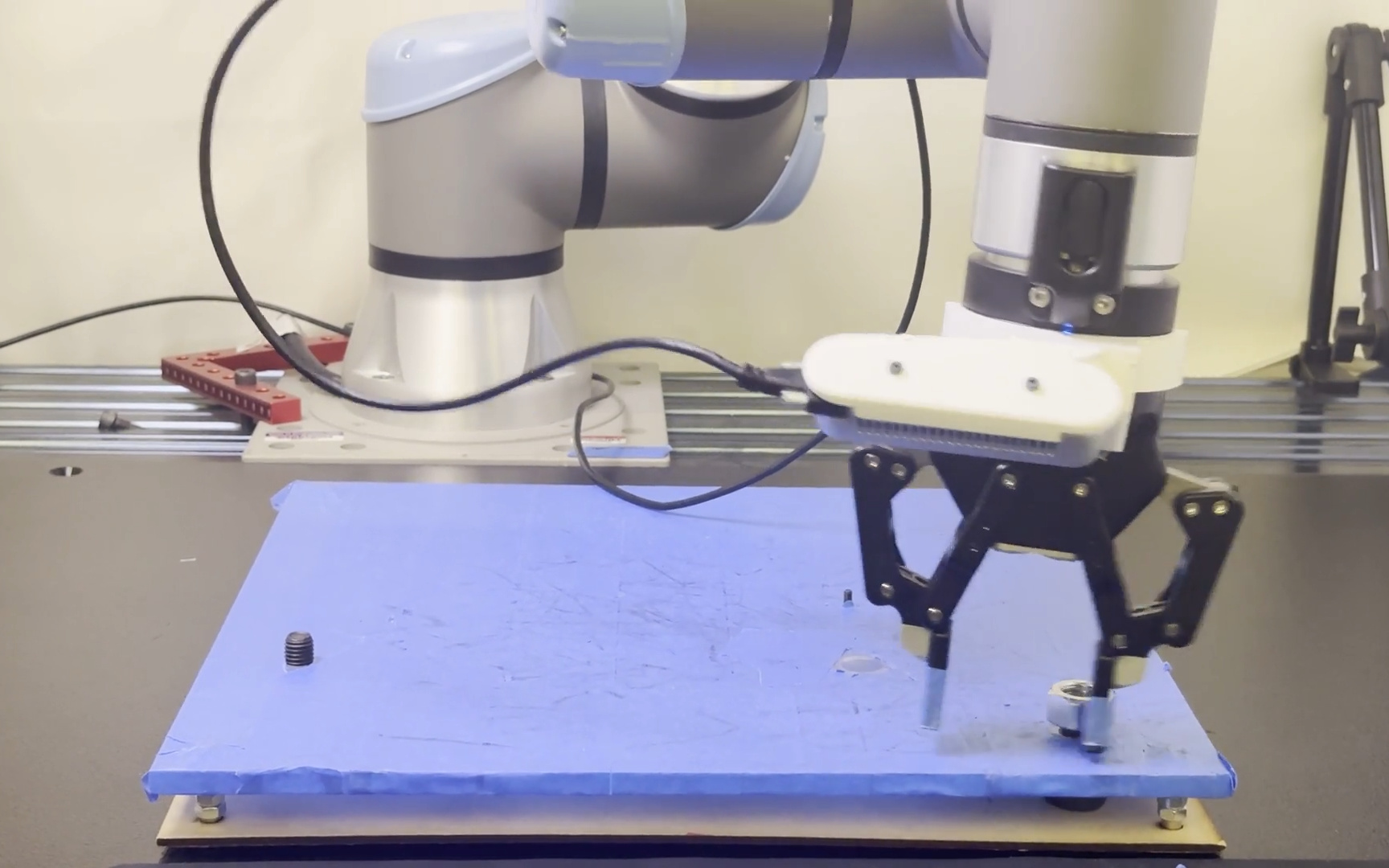}%
  \includegraphics[width=0.2\linewidth,trim={660bp 60bp 28bp 100bp},clip]{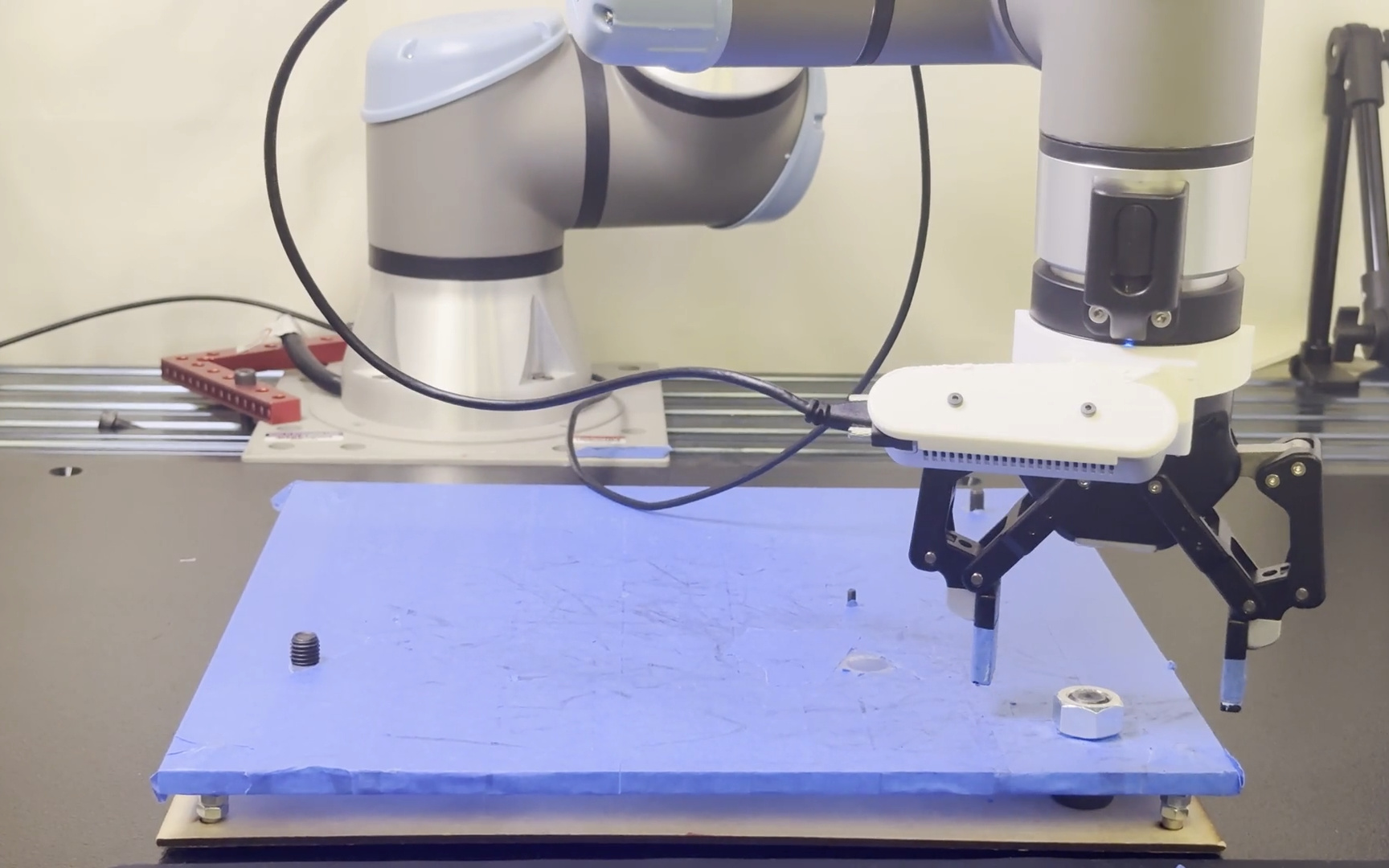}%
  \\[3pt]
  \includegraphics[width=0.2\linewidth,trim={0bp 0bp 3080bp 0bp},clip]{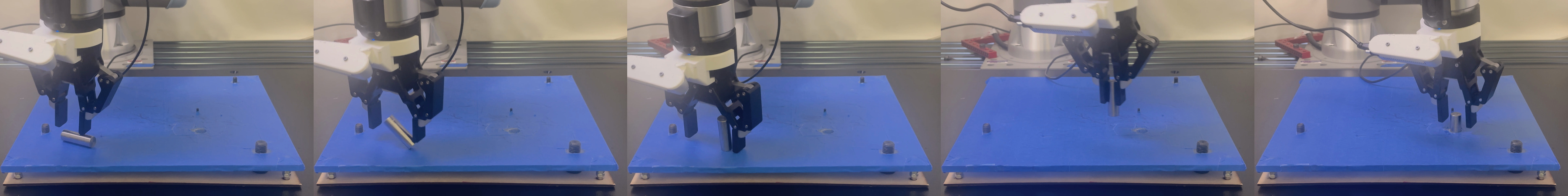}%
  \includegraphics[width=0.2\linewidth,trim={720bp 0bp 2360bp 0bp},clip]{FIGS/real-rod-filmstrip.jpg}%
  \includegraphics[width=0.2\linewidth,trim={1440bp 0bp 1640bp 0bp},clip]{FIGS/real-rod-filmstrip.jpg}%
  \includegraphics[width=0.2\linewidth,trim={2280bp 0bp 800bp 0bp},clip]{FIGS/real-rod-filmstrip.jpg}%
  \includegraphics[width=0.2\linewidth,trim={3000bp 0bp 80bp 0bp},clip]{FIGS/real-rod-filmstrip.jpg}%
  \\[3pt]
  \includegraphics[width=0.2\linewidth,trim={100bp 0bp 2980bp 0bp},clip]{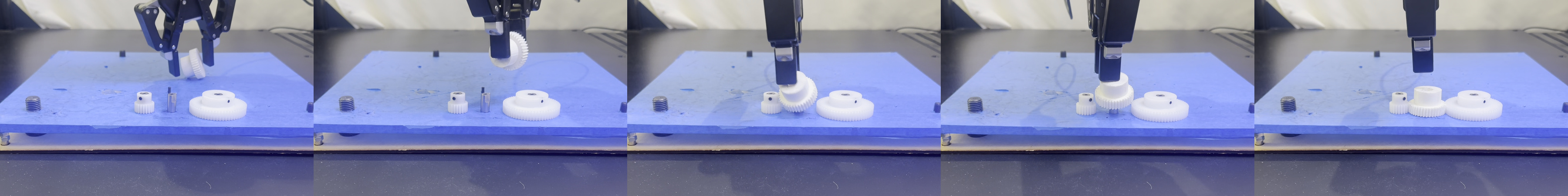}%
  \includegraphics[width=0.2\linewidth,trim={820bp 0bp 2260bp 0bp},clip]{FIGS/real-gear-filmstrip.jpg}%
  \includegraphics[width=0.2\linewidth,trim={1540bp 0bp 1540bp 0bp},clip]{FIGS/real-gear-filmstrip.jpg}%
  \includegraphics[width=0.2\linewidth,trim={2260bp 0bp 820bp 0bp},clip]{FIGS/real-gear-filmstrip.jpg}%
  \includegraphics[width=0.2\linewidth,trim={2980bp 0bp 100bp 0bp},clip]{FIGS/real-gear-filmstrip.jpg}%
  \caption{\textbf{Zero-shot RGB sim-to-real transfer.} Successful \taskNut{} (top), \taskRod{} (middle), and \taskGear{} (bottom) trajectories on UR5e hardware.}
  \label{fig:real_transfer}
\end{figure}

We study locomotion for a quadrupedal robot  traversing a variety of different terrains, some of which can be seen in Figure~\ref{fig:1_hero} (See Appendix \ref{appendix:locomotion_task_config} for a thorough description). We use a set of 13 terrain types, which span a wide range of difficulties and capabilities that the robot needs to obtain in order to traverse them successfully. These include terrains found in previous works, such as slopes, pits, gaps, stairs,  \citep{rudin2022advanced}, radiating beams, balance beams, stepping stones \citep{zhang2024learning}, and five harder procedurally generated terrains. These include navigation mazes, contoured terrains, and four variations of floating islands connected to each other via either narrow stepping stones or randomly arranged jump boxes \citep{hoeller2024anymal,cheng2024extreme,zhuang2023robot} at different heights. These terrains test both precise locomotion and navigation, requiring the robot to string together multiple agile maneuvers. Notably, we train a \textit{single} policy for all of these terrains from scratch, unlike many prior methods which train per-terrain specialist policies and then distill them into a single policy \citep{rudin2022advanced, zhang2024learning, hoeller2024anymal, zhuang2023robot}.
We use the Anymal-D robot \citep{hutter2017anymal} with system-identified actuators \citep{bjelonic2025towards}, to ensure realistic motions.

\subsection{Manipulation}
\begin{figure}[t]
  \centering
  \captionsetup[subfigure]{justification=centering}
  \begin{subfigure}[t]{0.32\linewidth}
    \centering
    \includegraphics[width=\linewidth]{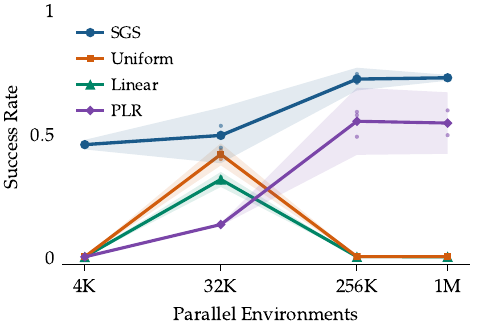}
    \caption{Multi-terrain locomotion}
    \label{fig:scaling_locomotion}
  \end{subfigure}\hfill
  \begin{subfigure}[t]{0.32\linewidth}
    \centering
    \includegraphics[width=\linewidth]{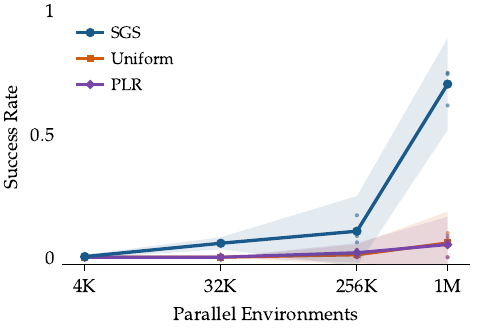}
    \caption{Franka \taskNut{}}
    \label{fig:scaling_manipulation}
  \end{subfigure}
  \hfill \begin{subfigure}[t]{0.32\linewidth} \centering \includegraphics[width=\linewidth]{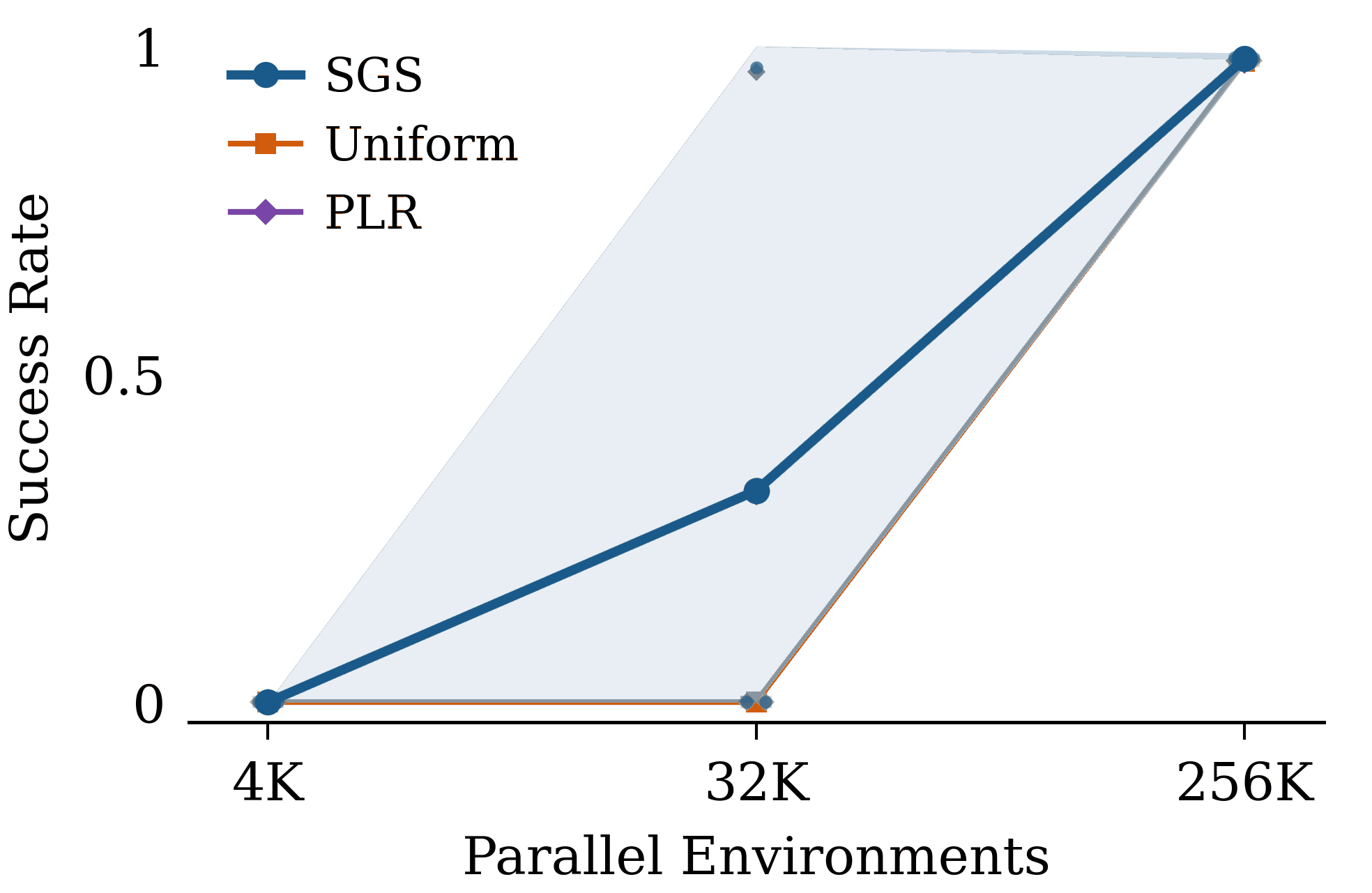} \caption{UR5e \taskRod{}} \label{fig:scaling_rod} \end{subfigure} \caption{Success rate versus the number of parallel training environments.
  Curves show means over three training seeds, dots show individual seeds,
  and shaded regions show 95\% confidence intervals.
  \methodname{} achieves higher mean success than the compared baselines at
  1M environments on locomotion and Franka \taskNut{}. On the simpler UR5e \taskRod{}, \methodname{} outperforms uniform sampling and PLR at smaller scale (32K), and all methods reach approximately 98\% success at high-enough scale (256K).}
\label{fig:4_5_locomotion_scaling}
\vspace{-1em}
\end{figure}

\begin{wrapfigure}{r}{0.3\textwidth}
  \centering
  \includegraphics[width=1.0\linewidth]{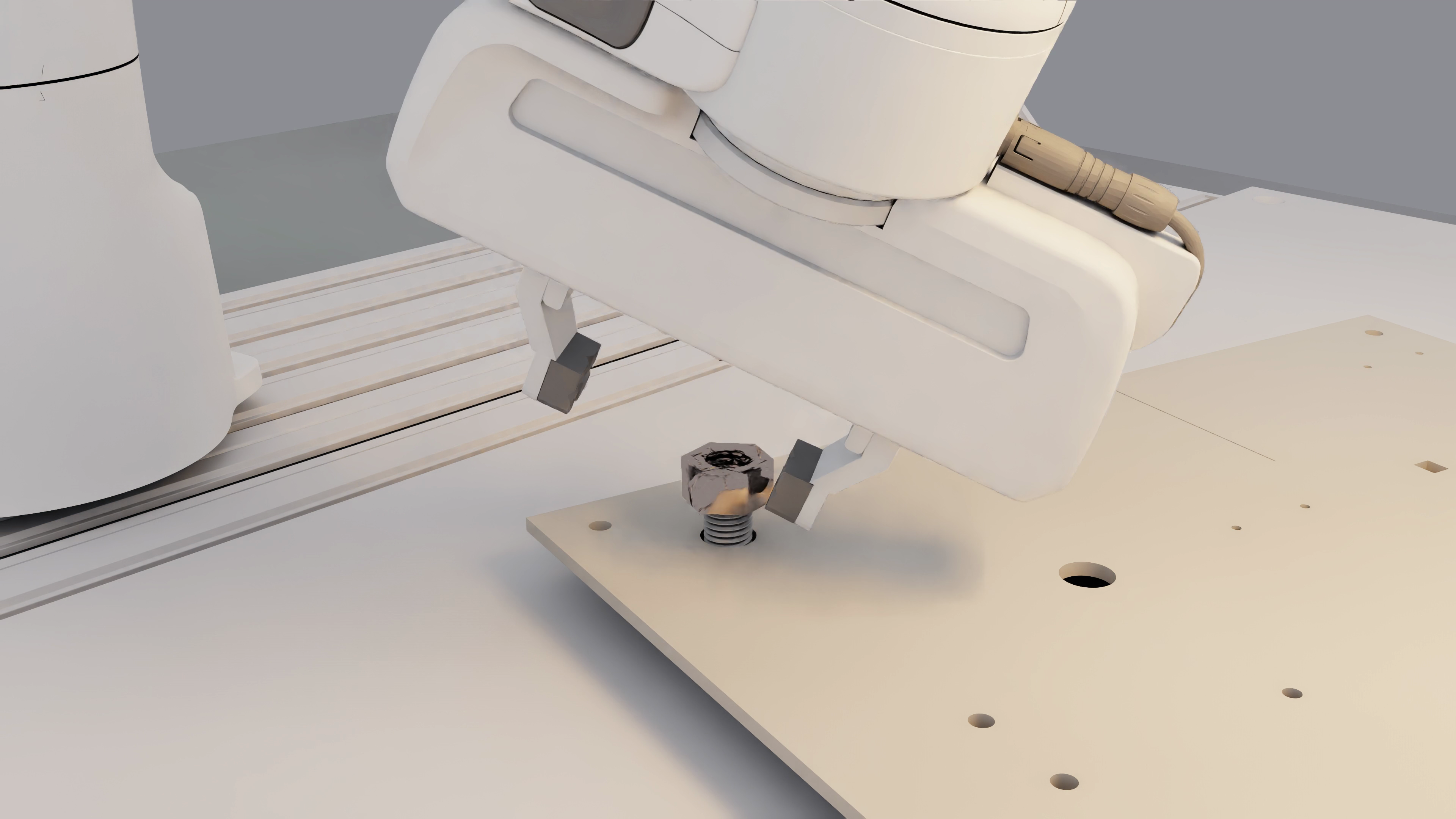}
  \caption{\taskNut{} with Franka evaluated in our scaling experiment.}
  \label{fig:franka_nut}
\end{wrapfigure}

We study contact-rich assembly on six UR5e tasks (Figure~\ref{fig:six_nist_tasks}, top row) and Franka \taskNut{} (Figure~\ref{fig:franka_nut}) from the NIST Assembly Taskboard~\citep{kimble2020benchmarking}. The UR5e tasks cover threading, precise insertion, connector mating, and gear alignment. The UR5e uses a Robotiq 2F-85 gripper and operational-space control; the Franka uses a parallel-jaw gripper and joint-position control.

Each task requires grasping and reorienting a part, transporting it to its mating geometry, and completing a precise, contact-rich assembly. In \taskNut{}, the robot threads an M16 nut onto an upright bolt through approximately $6.5$ revolutions. In \taskGear{}, it seats a medium gear on its shaft with its teeth meshing with the adjacent small and large gears. In \taskWaterproof{}, it inserts the connector into its socket until fully seated. In \taskBNC{}, it inserts the connector and rotates it to engage the lock. In \taskRod{}, it inserts a 16\,mm rod into the taskboard hole. In \taskPeg{}, it inserts a rectangular peg into its opening, requiring positional and rotational alignment. Appendices~\ref{appendix:manipulation_task_config} and~\ref{subsec:reward-manipulation} give task-specific reset and success definitions.

We use the same simple reward function and diverse simulator resets across all manipulation tasks. Our main scaling manipulation experiments use a gravity curriculum which we later found to be unnecessary; Appendix~\ref{app:gravity_ablation} describes its implementation and ablates its effect.

\subsection{\methodname~benefits from increased training scale.}
We compare the scaling behavior of \methodname~against three baselines: uniform sampling, hand-designed curricula (common in locomotion RL), and Prioritized Level Replay (PLR)~\citep{jiang2021prioritized}, a popular Unsupervised Environment Design method~\citep{dennis2020emergent}. Uniform sampling draws task configurations from $\rho$ at every reset. It has recently been shown to suffice for easier manipulation tasks~\citep{yin2026emergent} but has not been effective in multi-task locomotion. Locomotion instead commonly uses a ``linear'' curriculum~\citep{rudin2022advanced, rudin2022learning}, in which terrain difficulty is interpolated along a hand-designed heuristic such as stair height. Robots start at random levels and are promoted when they get within some distance of the goal or demoted when they make little progress. PLR samples the next task configuration using a ``learnability score'' plus a staleness bonus; we use the absolute GAE as the score.

We measure average success on Anymal-D locomotion and Franka Panda \taskNut{} across four training scales, up to 1M parallel environments. We report means over three training seeds per scale for both tasks. Evaluation resets are freshly sampled from the training distribution rather than reused from training. We hold the number of PPO mini-batches constant, so batch size scales with environment count, and fix all other PPO and optimizer hyperparameters. We train locomotion for 5K PPO iterations and \taskNut{} for 7.5K, with up to 32,768 and 16,384 environments per GPU, respectively, on L40S and H200 GPUs. Training takes about 36 hours for locomotion and 128 hours for \taskNut{}.

At 1M parallel environments, \methodname{} reaches 73\% mean success on multi-task locomotion versus 54\% for PLR (Figure~\ref{fig:4_5_locomotion_scaling}). On Franka Panda \taskNut{}, it reaches 70\%, versus 6\% for uniform sampling and 5\% for PLR. Notably, we see monotonic improvement in performance with increased scale, where other methods struggle to benefit or collapse. On the easier UR5e \taskRod{} task (Figure~\ref{fig:scaling_rod}), \methodname{} succeeds at 32K environments, where uniform sampling and PLR achieve zero success, while all methods reach about 98\% at 256K. \methodname{} also outperforms Sampling for Learnability (SFL)~\citep{rutherford2024no} on locomotion at 4K and 32K environments (Table~\ref{tab:sfl_comparison}). We report both SFL's best and final performance to show its late-training decline; implementation details are in Appendix~\ref{app:sfl_comparison}.

\begin{table}[htbp]
  \centering
  \caption{Locomotion success rate (mean $\pm$ 95\% confidence interval over three training seeds). SFL best is reported alongside final performance to show the effect of its late-training decline.}
  \label{tab:sfl_comparison}
  \begin{tabular}{lccc}
    \toprule
    Parallel environments & SFL final & SFL best & \methodname{} final \\
    \midrule
    4K & $0.255 \pm 0.051$ & $0.289 \pm 0.015$ & $0.456 \pm 0.019$ \\
    32K & $0.051 \pm 0.142$ & $0.421 \pm 0.032$ & $0.494 \pm 0.113$ \\
    \bottomrule
  \end{tabular}
\end{table}

\begin{figure}[t]
  \centering
  \includegraphics[width=\linewidth]{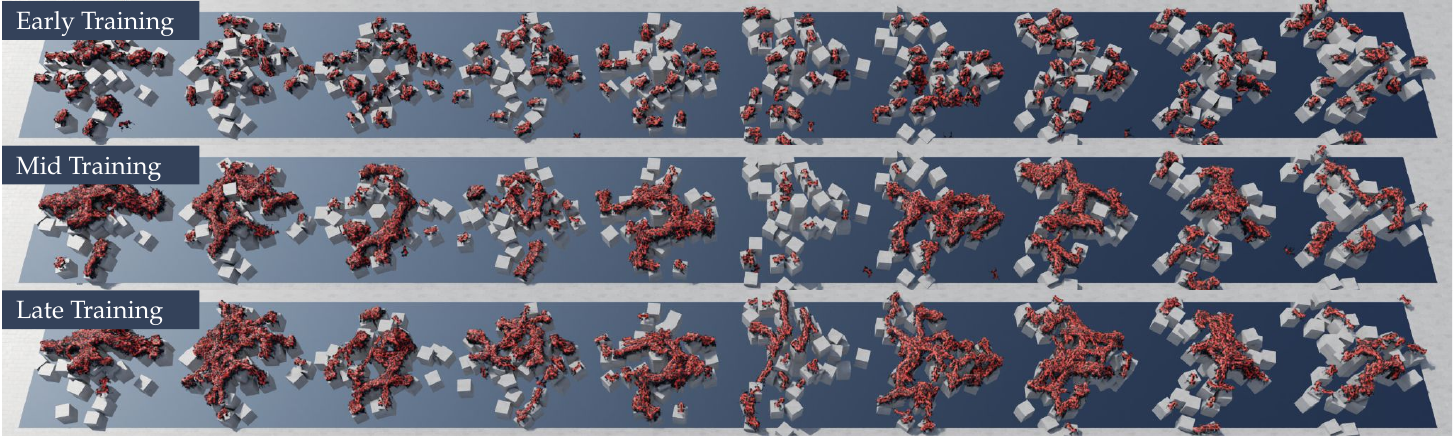}
  \caption{\textbf{Sampling of configurations under \methodname{} over training.}
  Early in training, \methodname{} samples uniformly over all task configurations. Halfway through training, \methodname{} focuses sampling mass on learning to jump across islands that are level to one another. In late training, \methodname{} samples configurations that include the top of the center pillar, which is at a significant height difference compared to the other islands.}
  \label{fig:locomotion_sgs_mass}
\end{figure}

\subsection{\methodname~unlocks new capabilities even at small scale.}
\label{sec:small_scale}
\methodname{}'s advantage over baselines emerges well before the saturation effects seen at large scale, as the UR5e \taskRod{} scaling comparison shows (Figure~\ref{fig:scaling_rod}). The same holds for multi-task locomotion: with only 4{,}096 parallel environments, \methodname{} achieves non-trivial success on many terrains while uniform sampling and other baselines fail to learn at all (\autoref{fig:4_6_locomotion_at_4k}). Even at small scale, \methodname{}'s adaptive sampling squeezes more useful learning signal under the same budget compared to uniform sampling.

\afterpage{\begin{figure}[H]
  \centering
  \includegraphics[width=\linewidth]{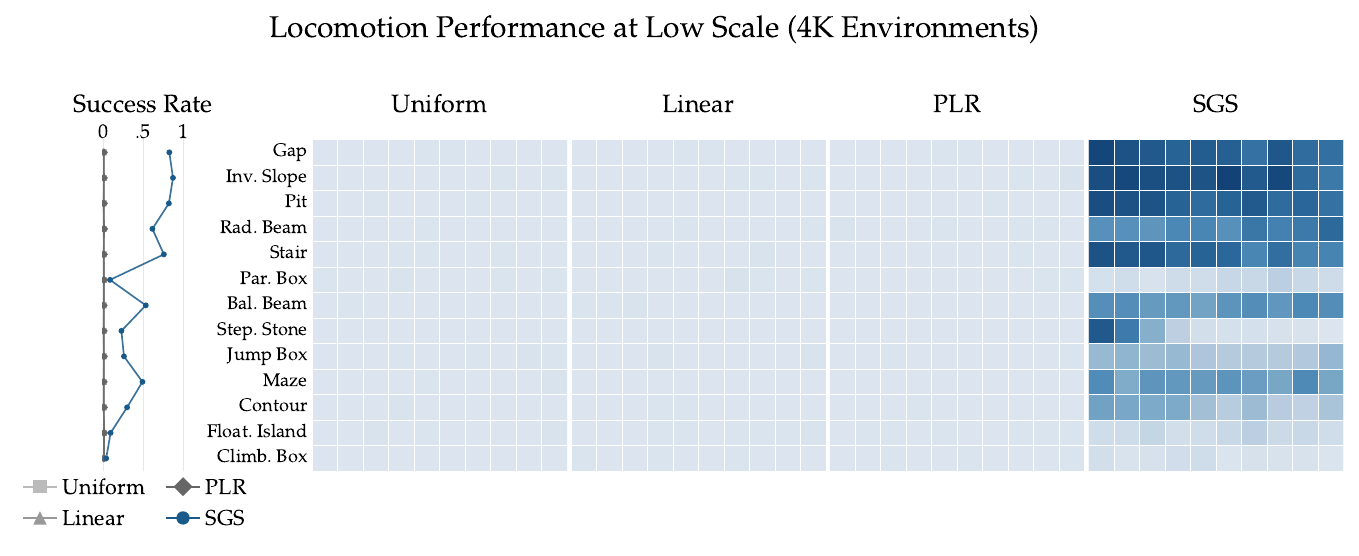}
  \caption{\textbf{Locomotion training at 4K environments.} Mean success rate at 4096 parallel environments. \methodname{} achieves non-zero success on a large portion of the terrains, while baselines do not learn at all at this scale.}
  \label{fig:4_6_locomotion_at_4k}
\end{figure}
}

\subsection{Zero-shot deployment on real hardware}
\label{sec:hardware}
For real-world transfer, we train additional UR5e assembly policies with \methodname{}. Appendix~\ref{app:additional_embodiments} reports their RL training setup and simulation comparisons.
We distill the state-based teachers into RGB policies with DAgger and transfer them zero-shot to UR5e hardware for \taskNut{}, \taskRod{}, and \taskGear{}. Distillation takes about two days on four L40S GPUs with 480 environments each. Table~\ref{tab:real_robot_results} reports teacher and RGB-student success in simulation alongside real-world success, first-try success (task completed on the first insertion attempt), and throughput (successes per minute of total evaluation time, including resets). \methodname{} policies exhibit robust retrying and dexterous non-prehensile manipulation, such as flipping the nut and pushing the gear against the shaft to reorient it (Figure~\ref{fig:real_transfer}). See Appendix~\ref{app:sim2real_transfer} for distillation, sim-to-real transfer, hardware, and evaluation details.

\begin{table}[htbp]
  \centering
  \caption{\textbf{Simulation and zero-shot hardware performance.} Simulation success is reported from the Reaching initial-state distribution. Hardware success counts and trial totals are shown in parentheses.}
  \label{tab:real_robot_results}
  \small
  \setlength{\tabcolsep}{3pt}
  \begin{tabular}{lccccc}
    \toprule
    & \multicolumn{2}{c}{Simulation success (\%)} & \multicolumn{3}{c}{Real hardware} \\
    \cmidrule(lr){2-3}\cmidrule(lr){4-6}
    Task & Teacher & RGB student & \shortstack{Success\\(\%)} & \shortstack{First-try\\success (\%)} & \shortstack{Throughput\\(successes/min)} \\
    \midrule
    \taskNut{} & 90.04 & 89.05 & 37.5 (18/48) & 12.5 & 0.23 \\
    \taskRod{} & 95.70 & 93.36 & 61 (30/49) & 28.57 & 0.71 \\
    \taskGear{} & 96.09 & 96.42 & 94 (47/50) & 76.0 & 3.08 \\
    \bottomrule
  \end{tabular}
  \par\smallskip
\end{table}

\section{Limitations}
\label{sec:limitations}
We find three primary limitations. First, \methodname~maintains success-rate estimates over a discrete set of $K$ cells; while $K=32{,}768$ is sufficient here, continuous or much larger task spaces would require a learned or hierarchical scoring representation. Second, \methodname~samples from a given distribution but does not construct it: we rely on reset-diversity methods such as OmniReset~\citep{yin2026emergent} for broad coverage, and closing the loop on \emph{which} configurations to make available remains an open problem. Third, a sim-to-real transfer gap remains, and closing it to match simulated success rates is left to future work.

\section{Conclusion}
\label{sec:conclusion}
We present \methodname{}, a simple success-based sampler that focuses training on task configurations with moderate success rates. Combined with diverse resets and large-scale RL, \methodname{} solves harder tasks than prior samplers under the same training budget, including quadruped locomotion and nut-and-bolt assembly. We distill the manipulation policies to RGB and demonstrate zero-shot transfer of three NIST board assembly tasks to UR5e hardware. Training policies  with~\methodname{} enables learning without as much hand-engineered reward specification nor manual curriculum engineering on challenging tasks compared to previous works. These results show that adaptively choosing what the policy practices is an important component of scaling robot RL, and suggest a path toward broader multi-task learning and general-purpose dexterous policies.

\label{sec:acknowledgements}
\acknowledgments{We thank the University of Washington's Hyak and Tillicum GPU-accelerated computing platforms, NVIDIA, and NSF ACCESS for the GPU compute that made it possible to run our experiments. This work used the DeltaAI system at the National Center for Supercomputing Applications through allocation CIS260956 from the Advanced Cyberinfrastructure Coordination Ecosystem: Services \& Support (ACCESS) program, supported by U.S. National Science Foundation grants \#2138259, \#2138286, \#2138307, \#2137603, and \#2138296 \citep{boerner2023access}. DeltaAI is supported by the National Science Foundation (award OAC 2320345) and the State of Illinois. Patrick Yin is supported by the National Science Foundation Graduate Research Fellowship Program. Abhishek Gupta acknowledges support from Toyota Research Institute under the University 3.0 Research Program. We thank Nam Pho for administering the ACCESS allocation, Andrey Kolobov at Microsoft Research for lending us a NIST task board, and Emma Romig for help building a prototype NIST board at UW. We thank members of the University of Washington's Robot Learning and WEIRD Lab for useful discussions and feedback.
}

\clearpage
\bibliography{example}  %

\clearpage
\appendix
\section{Task Specification Details}

\subsection{Locomotion}
\label{appendix:locomotion_task_config}

For locomotion, we define $\rho$ as a discrete categorical distribution $\rho = (T, L, s_0, g)$ over terrain type $T$, terrain level $L$, initial state $s_0$, and goal state $g$. We use 2 seeds per terrain type, 10 difficulty levels, 20 start states, and 20 goal states, yielding $104{,}000$ task configurations.

Terrains from previous works typically have a linear structure where difficulty is determined by parameters such as slope or step height, linearly interpolated from easy to hard. For our procedurally generated terrains, a clear ordering in terms of difficulty may not exist, so the difficulty axis can be interpreted as different seeds.

The terrains used in this paper are shown in the middle and bottom rows of Figure~\ref{fig:locomotion_tasks}, with the exception of \texttt{Inverted Slope} which we omit due to space constraints.

\subsection{Manipulation}
\label{appendix:manipulation_task_config}

We train a separate policy for each manipulation task, whereas locomotion uses a single policy trained jointly across terrains. For manipulation, we define $\rho$ as a discrete categorical distribution $\rho = (c, s_0, g)$ over reset configurations $c$, the initial state $s_0$ they emit, and the goal $g$. Here $g$ is fixed across all task configurations to the fully-assembled pose in the fixed-asset frame. The current single-task setup pre-collects a buffer of $32{,}768$ collision-checked configurations balanced across three reset strategies; at training time, every episode reset draws a state from this buffer. We describe each of these reset strategies below:

\paragraph{Near-Goal.} This reset strategy places the manipulated object, e.g. the nut in \taskNut{}, along its assembly path, e.g. the bolt it threads on. We position the end-effector around a task-specific grasp point using inverse kinematics, randomizing its position, orientation, and gripper aperture. For \taskRod{}, the path follows the hole axis and the current fractional range is $[0,1.1]$, where zero is fully seated and values above one include slightly disengaged starts. Other tasks use their own assembly profiles and fractional ranges. For instance, threaded assembly couples axial translation with rotation.

\paragraph{Stable Grasp.} This reset strategy places the object above the assembly target, at a higher $z$ position than Near-Goal, and uses the same task-specific grasp setup with a randomized gripper aperture.

\paragraph{Reaching.} This reset strategy samples the object's pose anywhere in the workspace, then positions the end-effector to be aligned the object's grasp frame using inverse kinematics. The gripper aperture is randomized, so the reset does not guarantee that the object is already secured in a stable grasp. This can lead to the object falling anywhere in the workspace, requiring the robot to go fetch the object and then complete the task.

These reset families are used in the ablation in Appendix~\ref{app:reset_coverage}. Franka \taskNut{} uses the same reset families and reset-buffer construction as UR5e.

\paragraph{Embodiment-specific simulation and control.}
Franka uses joint-position control, whereas UR5e uses operational-space control (OSC). Both use a physics time step of 5\,ms (200\,Hz), but their policy frequencies differ: Franka runs at 25\,Hz with eight physics steps per action, while UR5e runs at 10\,Hz with 20 physics steps per action. Thus, Franka requires 2.5 times fewer physics steps per collected policy transition, contributing to its faster training. The reported Franka \taskNut{} runs take approximately 128 hours, compared with 335 hours for UR5e \taskNut{}.

\subsection{Observations, Actions, and Rewards}
\label{appendix:reward_specification}

We describe the observation and action spaces for each embodiment, followed by the reward definitions and weights.

\subsubsection{Locomotion}
\label{subsec:reward-locomotion}

\paragraph{Observations and actions.}
The locomotion policy observes base linear and angular velocity, projected gravity, joint positions and velocities, the previous action, the goal command, and a local terrain height scan. Its 12-dimensional action specifies joint-position offsets from the robot's nominal pose, which are tracked by the joint controllers.

\paragraph{Rewards.}
Since we train a single policy across all terrain types, we use the same reward function across all locomotion tasks. It combines boolean terminal success and failure terms, which contain the bulk of the reward signal, with per-step terms for motion, posture, contacts, and forward alignment:
\begin{equation}
\label{eq:reward_form_locomotion}
  r(s_t, a_t) \;=\;
  \underbrace{\lambda_1 \cdot r_{\text{succ}} \,+\, \lambda_2 \cdot r_{\text{fail}}}_{\text{terminal terms}}
  \;+\;
  \underbrace{\lambda_3 \cdot r_{\text{work}} \,+\, \lambda_4 \cdot r_{\text{post}} \,+\, \lambda_5 \cdot r_{\text{td}} \,+\, \lambda_6 \cdot r_{\text{undc}} \,+\, \lambda_7 \cdot r_{\text{fwd}}}_{\text{per-step terms}}.
\end{equation}

\paragraph{Term definitions.}
\begin{align*}
r_{\text{succ}}(s_t) &= \mathbf{1}[\text{command-success at }t] && \text{(terminal success)} \\
r_{\text{fail}}(s_t) &= \mathbf{1}[\text{drop or base-contact termination at }t] && \text{(terminal failure)} \\
r_{\text{work}}(s_t,a_t) &= \,\bigl\lVert\tau_t \odot \dot q_t\bigr\rVert_1 && \text{(L1 mechanical power)} \\
r_{\text{post}}(s_t) &= \,\bigl\lVert q_t - q_{\text{nom}}\bigr\rVert_1 && \text{(L1 posture deviation)} \\
r_{\text{td}}(s_t) &= \!\!\sum_{f\in\text{feet}}\!\!\mathbf{1}[\text{TD}_f(t)]\,\bigl\lVert v_{f,z}(t)\bigr\rVert_2^2 && \text{(squared touchdown velocity)} \\
r_{\text{undc}}(s_t) &= \!\!\sum_{i\in\mathcal{B}_{\text{non-foot}}}\!\!\mathbf{1}\!\bigl[\lVert F_i(t)\rVert_2>F_{\min}\bigr] && \text{(undesired hard contacts)} \\
r_{\text{fwd}}(s_t) &= \max\!\bigl(0,\,\bigl\langle \hat v_t^{(b)},\, \hat g_t^{(b)}\bigr\rangle\bigr) && \text{(forward-aligned motion, clamped)}
\end{align*}

$r_{\text{fwd}}(s_t)$ is used to promote the robot walking forwards rather than towards the goal but facing backwards. This reward is deactivated within the first 500 steps of training. 

\paragraph{Weights.}
\begin{equation*}
\begin{aligned}
\lambda_1\,(\text{succ}) &= 50, &\quad
\lambda_2\,(\text{fail}) &= -50, \\
\lambda_3\,(\text{work}) &= -10^{-3}, &\quad
\lambda_4\,(\text{post}) &= -5\!\times\!10^{-2}, \\
\lambda_5\,(\text{td}) &= -2.5\!\times\!10^{-1}, &\quad
\lambda_6\,(\text{undc}) &= -5\!\times\!10^{-2}, \\
\lambda_7\,(\text{fwd}) &= 0.3.
\end{aligned}
\end{equation*}

\subsubsection{Manipulation}
\label{subsec:reward-manipulation}

\paragraph{Observations and actions.}
The teacher observes joint positions, end-effector pose, the previous action, and a scene point cloud of the gripper, held object, and board expressed in the gripper frame. The action comprises three Cartesian translation increments, three axis-angle rotation increments, and one binary gripper open/close command. The arm commands are executed by an operational-space controller. Appendix~\ref{app:distillation} describes distillation into an RGB student for deployment.

\paragraph{Rewards.} The UR5e and Franka assembly tasks share the same reward function, combining assembly success with safety and motion regularization:
\begin{equation}
\label{eq:reward_form_manipulation}
  r(s_t, a_t) \;=\;
  \underbrace{r_{\text{succ}}}_{r_{\text{task}}}
  \;+\;
  \underbrace{r_{\text{fail}} \,+\, r_{\text{act}} \,+\, r_{\Delta a} \,+\, r_{\dot q}}_{r_{\text{reg}}}.
\end{equation}

\paragraph{Assembly success and safety regularization.} Let $S(s_t)$ denote the task-specific assembly-success predicate and $F(s_t)$ indicate termination from abnormal joint velocities or the held object leaving the configured workspace bounds. The task reward and terminal safety penalty are
\begin{equation*}
r_{\text{succ}}(s_t)=\lambda_{\text{succ}}\mathbf{1}[S(s_t)],
\qquad
r_{\text{fail}}(s_t)=-\lambda_{\text{fail}}\mathbf{1}[F(s_t)].
\end{equation*}
The failure penalty is a safety regularizer applied at termination. The abnormal-velocity condition uses twice the configured joint-velocity limits. Success is determined by orientation, lateral-position, and insertion-depth gates using the task's object geometry; the exact depth bounds and reference points depend on the task.

\paragraph{Per-step regularizers.} Define $C(x)=\min(x,10^4)$ for nonnegative $x$. The implementation clamps each squared-norm penalty:
\begin{align*}
r_{\text{act}}(s_t,a_t) &= -\lambda_{\text{act}} C\!\left(\lVert a_t\rVert_2^2\right) && \text{(action magnitude)} \\
r_{\Delta a}(s_t,a_t) &= -\lambda_{\Delta a} C\!\left(\lVert a_t-a_{t-1}\rVert_2^2\right) && \text{(action rate)} \\
r_{\dot q}(s_t,a_t) &= -\lambda_{\dot q} C\!\left(\lVert\dot q_{t,\mathrm{arm}}\rVert_2^2\right) && \text{(arm-joint velocity)}.
\end{align*}
Here $\dot q_{t,\mathrm{arm}}$ corresponds to the arm joints of the respective robot.

\paragraph{Configured weights.}
\begin{equation*}
\begin{aligned}
\lambda_{\text{succ}} &= 100, &\quad \lambda_{\text{fail}} &= 1, &\quad \lambda_{\text{act}} &= 10^{-5}, \\
\lambda_{\Delta a} &= 10^{-4}, &\quad \lambda_{\dot q} &= 10^{-4}.
\end{aligned}
\end{equation*}

\section{Additional Experimental Results}
\label{app:additional_experiments}

\subsection{Comparison with Sampling for Learnability}
\label{app:sfl_comparison}
Table~\ref{tab:sfl_comparison} in the main text compares SFL \citep{rutherford2024no} with \methodname{} on locomotion. We use SFL's default hyperparameters and report means and 95\% confidence intervals over three training seeds. SFL's final performance is lower than \methodname{} at both scales. At 32K, its best checkpoint comes close, but still underperforms SGS. Notably, performance significantly degrades as training continues, most noticeably at larger scales, whereas SGS monotonically improves during training.

\subsection{\methodname{} Hyperparameters and Configuration Density}
\label{app:sgs_sensitivity}
\paragraph{Sampler implementation details.} For each task configuration, we store a window of the latest $H=100$ Boolean outcomes in a circular buffer. Our sampling floors the kernel weight before forming logits and enforces a minimum temperature of one:
\begin{equation*}
\widetilde w_i=\max(w_i,\epsilon),\qquad
\ell_i=\log(\widetilde w_i+\epsilon),\qquad
T_{\mathrm{eff}}=\max(T,1).
\end{equation*} 

\paragraph{Sensitivity experiments.}
To evaluate the sensitivity of~\methodname{} to each hyperparameter, we vary each \methodname{} hyperparameter one at a time on locomotion at 32K parallel environments, from the nominal setting used during training (in bold) using two training seeds per setting (Table~\ref{tab:sgs_sensitivity}). Across the tested hyperparameter settings, the mean success ranges from 0.39 to 0.53, compared with 0.49 for the nominal setting, showing that~\methodname{} is not very sensitive to hyperparameters, although it does benefit from a small hyperparameter search. We also compare the Beta kernel against a Normal kernel. Both Normal and Beta kernels achieve similar mean success at 4K environments (0.45 and 0.46, respectively over three training seeds). 

We also evaluate how sensitive~\methodname{} is to the task configuration density. For task configuration density, each terrain tile uses the product of $p$ spawn and $p$ goal patches. Increasing $p$ from 10 to 20 (i.e. from $52,000$ task configurations to $104,000$) increases success from 0.38 to 0.49, while $p=40$ (i.e. $208,000$ task configurations) yields 0.48 at 32K environments. This shows that the sampler benefits from having higher diversity up to a certain extent, after which performance plateaus.

To find the hyperparameter values used during training, we ran a small grid search over different values. For manipulation, a sweep over $\kappa\in\{1,5,10\}$ and $t\in\{0.33,0.5,0.66\}$ with $T=2$ at 32K parallel environments yielded $\kappa=1$ and $t=0.5$ as the best hyperparameters, which performed best by a small margin. Note that we typically do not modify $T$, since it is tied to $\kappa$, which we do sweep over.

\begin{table}[htbp]
  \centering
  \small
  \setlength{\tabcolsep}{4pt}
  \caption{\textbf{Locomotion sensitivity experiments.} Each entry gives the setting followed by its success rate in parentheses. Bold denotes the nominal values used during training. Hyperparameter experiments use 32K environments and two training seeds per setting. Kernel experiments use 4K environments and three training seeds. Patch-density experiments use 32K environments.}
  \label{tab:sgs_sensitivity}
  \begin{tabular}{@{}lccccc@{}}
    \toprule
    Target $t$ & .33\,(.50) & .50\,(.44) & \textbf{.66\,(.49)} & .80\,(.42) & .90\,(.52) \\
    $\kappa$ & 1\,(.44) & 2\,(.47) & \textbf{5\,(.49)} & 10\,(.46) & \\
    Temperature $T$ & \textbf{2\,(.49)} & 4\,(.53) & & & \\
    History $H$ & 25\,(.39) & \textbf{100\,(.49)} & 400\,(.47) & & \\
    Floor $\epsilon$ & $10^{-8}$\,(.49) & $\mathbf{10^{-4}}$\,\textbf{(.46)} & $10^{-2}$\,(.44) & & \\
    \midrule
    Kernel & Normal\,(.45) & \textbf{Beta\,(.46)} & & & \\
    \midrule
    Patches $p$ & 10\,(.38) & \textbf{20\,(.49)} & 40\,(.48) & & \\
    \bottomrule
  \end{tabular}
\end{table}

\subsection{Sensitivity to Reset-Distribution Coverage}
\label{app:reset_coverage}
We evaluate the sensitivity of~\methodname{} to coverage of resets in terms of entire reset strategies. We retrain UR5e \taskRod{} using buffers constructed from two of the three reset distributions, with three training seeds per setting (Table~\ref{tab:ur_rod_ablation}). Removing Near-Goal prevents successful learning in the tested runs. Removing Reaching leaves success on the retained distributions essentially unchanged. Removing Stable Grasp reduces mean success, with one of the three seeds collapsing while the other two retain high performance. These results highlight the importance of near-goal coverage in this reset ablation.

\begin{table}[htbp]
  \centering
  \caption{\textbf{\taskRod{} reset-distribution ablation.} Mean success (\%) per reset distribution over three training seeds. Dashes indicate omitted reset distributions.}
  \label{tab:ur_rod_ablation}
  \begin{tabular}{lccc}
    \toprule
    Training buffer & Stable Grasp & Reaching & Near-Goal \\
    \midrule
    All three & 99.5 & 96.0 & 100.0 \\
    Without Stable Grasp & --- & 63.8 & 81.3 \\
    Without Reaching & 99.4 & --- & 100.0 \\
    Without Near-Goal & 0.0 & 0.0 & --- \\
    \bottomrule
  \end{tabular}
\end{table}

\subsection{Gravity-Curriculum Ablation}
\label{app:gravity_ablation}
Our main manipulation experiments, including the Franka scaling results, use a gravity curriculum. Training begins in zero gravity, and gravity increases toward its standard value once success exceeds 80\% at the current level. After completing the scaling experiments, we trained \methodname{} on Franka \taskNut{} at 1M environments without this curriculum, using one training seed. On the Reaching reset distribution, this run achieves 69.4\% success, compared with 70.2\% with the curriculum (Table~\ref{tab:gravity_ablation}), suggesting that the gravity curriculum is unnecessary for strong performance in this setting.

\begin{table}[htbp]
  \centering
  \caption{\textbf{Gravity-curriculum ablation at 1M parallel environments.} Success on the hardest \taskNut{} reset strategy. Values are success rates $\pm$ 95\% confidence intervals over 512 evaluation episodes, not across training seeds. The no-curriculum ablation uses one training seed.}
  \label{tab:gravity_ablation}
  \begin{tabular}{lc}
    \toprule
    Training setting & Success rate \\
    \midrule
    \methodname{} with gravity curriculum & $0.702 \pm 0.039$ \\
    \methodname{} without gravity curriculum & $0.694 \pm 0.040$ \\
    \bottomrule
  \end{tabular}
\end{table}

\subsection{Additional UR5e Assembly Tasks for Transfer}
\label{app:additional_embodiments}

We train additional UR5e assembly policies to support real-world transfer. Table~\ref{tab:six_task_manipulation} reports success from the Reaching initial-state distribution under full gravity ($1g$) across the six tasks in Figure~\ref{fig:six_nist_tasks}. Across samplers, \taskNut{} is trained with 256K parallel environments and the other five tasks with 64K. These results are separate from the Franka \taskNut{} and UR5e \taskRod{} scaling experiments.

\begin{table}[htbp]
  \centering
  \caption{\textbf{UR5e evaluation success (\%).} Policies are evaluated from the Reaching initial-state distribution.}
  \label{tab:six_task_manipulation}
  \setlength{\tabcolsep}{4pt} %
  \begin{tabular}{llrrrr}
    \toprule
    Task type & Task & \makecell[r]{Training\\envs.} & Uniform & PLR & \methodname{} \\
    \midrule
    \multirow{5}{*}{Insertion}
      & \taskGear{} & 64K & 94.24 & 95.51 & 96.09 \\
      & \taskWaterproof{} & 64K & 95.31 & 95.90 & 95.80 \\
      & \taskPeg{} & 64K & 92.09 & 84.18 & 92.77 \\
      & \taskRod{} & 64K & 95.90 & 95.41 & 95.70 \\
      & \taskBNC{} & 64K & 83.40 & 81.45 & 84.08 \\
    \midrule
    \makecell[l]{Gear alignment\\and threading} & \taskNut{} & 256K & 48.83 & 59.77 & 90.04 \\
    \bottomrule
  \end{tabular}
\end{table}

\paragraph{Comparison with Franka.}
The higher UR5e \taskNut{} success should be interpreted in the context of its different control setup. UR5e uses operational-space control (OSC), allowing the policy to command Cartesian end-effector motion directly, whereas Franka uses joint-position control and must learn to coordinate its joints to produce the required motion. We attribute much of the performance difference to this control interface. The embodiments and grippers also differ, and policies act at 10\,Hz on UR5e versus 25\,Hz on Franka.

\paragraph{Effect of task difficulty.}
\methodname{} provides its largest gain on \taskNut{}: 90.04\% success versus 48.83\% for uniform sampling and 59.77\% for PLR. On the five simpler insertion tasks, \methodname{} performs comparably to uniform sampling. We hypothesize that less forgiving contact dynamics and longer horizons make nut-and-bolt assembly benefit more from concentrating training on configurations with moderate success rates. For simpler insertion tasks, sufficient scale can make uniform sampling effective, although adaptive sampling can improve efficiency: in the separate \taskRod{} scaling experiment, \methodname{} succeeds at 32K environments while uniform sampling and PLR achieve zero success, and all methods reach about 98\% at 256K (Figure~\ref{fig:scaling_rod}).

\section{Sim-to-Real Transfer and Evaluation}
\label{app:sim2real_transfer}
We build on the transfer pipeline of OmniReset~\citep[Appendix A.3]{yin2026emergent}. Below, we describe the hardware changes, simulation settings, and teacher--student design used for our NIST-board tasks, followed by transfer observations and the real-world evaluation protocol.

\subsection{Hardware and Control Setup}
\label{app:hardware_modifications}
We use a UR5e arm with a Robotiq 2F-85 gripper and follow OmniReset's relative Cartesian operational-space control interface~\citep[Appendices A.3.5--A.3.6]{yin2026emergent}. Our camera setup uses one third-person camera and one wrist-mounted camera, rather than OmniReset's three cameras. We describe the point-cloud teacher and RGB student in Appendix~\ref{app:distillation}.

\paragraph{Board surface.}
The NIST board is made of polished acrylic. When the wrist camera faces the board directly, the robot observes its own reflection, which is out of distribution. The low surface friction also impairs policies that use the board to reorient the object. For these reasons, we covered the board with tape during evaluation.

\subsection{Simulation Modeling and Domain Randomization}
\label{app:contact_friction}

\paragraph{Domain randomization.}
\label{app:domain_randomization}

We follow OmniReset's domain-randomization setup~\citep{yin2026emergent}, adding small Gaussian noise to the point-cloud observations and randomizing object masses around their measured real-world values.

\paragraph{Gripper pad friction.} We use a fixed pad friction of $\mu=2$ with the \texttt{max} friction-combine mode to reduce unrealistic grasps while retaining stable training.

OmniReset~\citep{yin2026emergent} gave the Robotiq 2F-85 pads a friction coefficient of $\mu = 100$ with the max combine mode. This makes grasping forgiving and training easy, but we found it to be a significant source of sim-to-real gap. Any grasp that applies normal force succeeds in simulation, because the object stays frozen between the pads even when the grasp is physically implausible (Figure 9).

\begin{figure}[!htbp]
    \centering
    \includegraphics[width=0.5\linewidth]{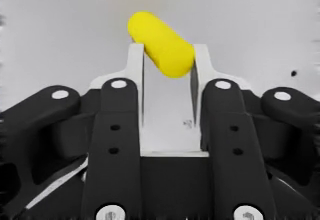}
    \caption{Example of an unrealistic grasp produced by 100$\mu$ gripper pads}
    \label{fig:pad_friction}
\end{figure}

Lowering the friction removes these grasps, but in our initial experiments policies trained from scratch at low pad friction failed to learn at all. To find the lowest friction that still trains, we trained \taskRod{} from scratch at fixed pad frictions of $\mu = 0.5$, $1$, $1.5$, and $2$, keeping the max combine mode and all other settings identical (Figure 10). Only $\mu = 2$ trained stably. We therefore use $\mu = 2$ for all training runs.(Figure~\ref{fig:pad_friction}).

\begin{figure}[!htbp]
    \centering
    \includegraphics[width=0.6\linewidth]{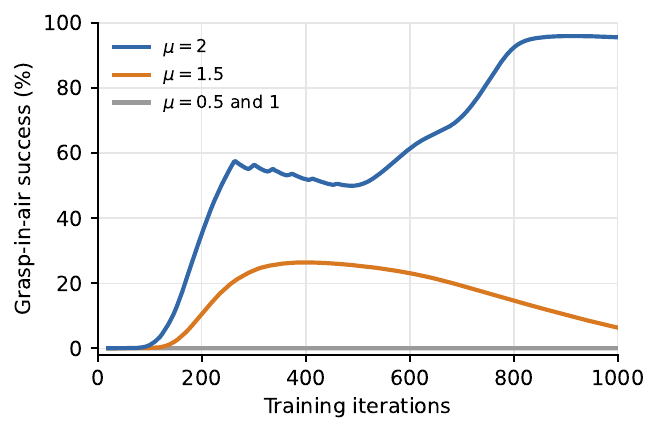}
    \caption{\textbf{Training \taskRod{} from scratch at pad friction $\mu = 0.5$, $1$, $1.5$, and $2$.} Grasp-in-air success rate over training (\texttt{max} combine, pad friction fixed for the whole run)}
    \label{fig:pad_friction_scratch}
\end{figure}

\subsection{Teacher--Student Distillation}
\label{app:distillation}

For UR5e transfer, \methodname{} trains a point-cloud teacher in simulation and distills it into an RGB student for deployment. The teacher actor observes arm joint positions, the previous action, and a scene point cloud expressed in the gripper frame; the critic additionally receives privileged state, including joint velocities and the sampled domain-randomization parameters. We use a point-cloud teacher rather than one that observes object poses because geometric observations narrow the gap between what the teacher and the RGB-based student can perceive, which makes the teacher's behavior easier to imitate from images. The student observes side and wrist RGB images, each encoded by a separate ResNet-18, together with arm joint positions, end-effector pose, and the previous action. OmniReset collects a dataset and trains the student with behavioral cloning~\citep{yin2026emergent}. We instead use DAgger for faster training and a simpler workflow, eliminating the separate dataset-collection step. During distillation, we use OmniReset's visual randomization and image augmentation~\citep[Table~2]{yin2026emergent}. Only the RGB student is run on hardware, without real-world data or fine-tuning.

\begin{figure}[htbp]
  \centering
  \includegraphics[width=0.72\linewidth]{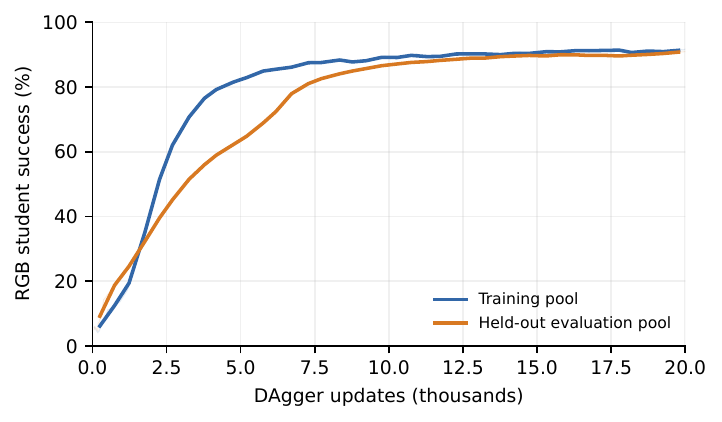}
  \caption{\textbf{Training and held-out evaluation during RGB distillation on \taskNut{}.} Evaluation success lags training success early on and catches up with continued training; curves show 500-update averages from one run.}
  \label{fig:dagger_train_eval}
\end{figure}

\paragraph{Online DAgger and held-out evaluation.}
Each iteration collects one step across all environments, updates the student toward the teacher's actions, and discards the data. This limits GPU memory use, a bottleneck when simulation and rendering run alongside training. However, training success can be misleading: updates from the immediately preceding step can guide the student's next action toward the teacher, helping it solve a task it cannot yet perform without these updates. We therefore reserve 6.67\% of environments for evaluation and exclude their data from gradient updates. Success in these held-out environments lags training success, revealing the need for longer distillation (Figure~\ref{fig:dagger_train_eval}).

\paragraph{Symmetry-aware resets and success criteria.} We randomize resets over each object's equivalent roll, pitch, and yaw angles. Discrete angles are sampled uniformly; continuously symmetric angles are sampled uniformly over $[0,360^\circ)$. The success reward accepts any of the same equivalent goal orientations. These rotations are defined in the object-local frame relative to the nominal configuration. 

Using the same symmetries for resets and rewards avoids favoring an orientation that the RGB-based student cannot distinguish from an equivalent one. Point-cloud observations alone do not remove this ambiguity if resets or rewards still impose a preferred orientation. Such preferences can also induce unnecessary reorientation behaviors that transfer poorly, as illustrated by the nut-flipping failure in Appendix~\ref{app:task_specific_gaps}.

On OmniReset peg insertion, the distilled RGB student achieves approximately 55\% success~\citep[Table~1]{yin2026emergent}. Using a point-cloud teacher with symmetric resets and rewards increases student success to 98--99\%. Each grasp reset uniformly samples one of eight equivalent object orientations, and the reward accepts successful insertion in any of them. This reduces the teacher--student observability gap by removing orientation distinctions unavailable to the RGB student, making the teacher easier to imitate.

\begin{table}[htbp]
\centering
\caption{\textbf{RGB-student success in simulation on OmniReset peg insertion.} The pose-based baseline is reported in OmniReset~\citep[Table~1]{yin2026emergent}; the symmetry-aware point-cloud teacher yields 98\% success after distillation.}
\label{tab:symmetry_dagger}
\begin{tabular}{lc}
\toprule
Teacher configuration & RGB student success (\%) \\
\midrule
Pose-based (OmniReset) & 55.45 \\
Eight-orientation, point-cloud & 98 \\
\bottomrule
\end{tabular}
\end{table}

\subsection{Transfer Observations and Policy Behaviors}
\label{app:task_specific_gaps}

A consistent lesson across tasks was that policies which commit to a clean grasp and insert directly transfer better than ones that exploit simulator contact dynamics. For \taskRod{}, adding 1 mm per-step Gaussian noise to the point-cloud observation further helped by encouraging the policy to re-aim after a miss and wiggle the rod when it was not fully seated. For \taskNut{}, the policy repeatedly dropped the nut onto the bolt instead of seating it, flipping the nut each time. We speculate this was due to the teacher reward, which only counted one upright nut orientation for success despite the nut being symmetric. Removing this constraint and using appropriately tuned action scaling on the 6-DOF end-effector control space produced a policy that completed the task.

\subsection{Real-World Evaluation Protocol}
\label{app:real_world_details}
Figure~\ref{fig:real_transfer} shows hardware trajectories, and Table~\ref{tab:real_robot_results} reports task-level results. We follow the evaluation protocol and metric definitions of OmniReset~\citep[Appendix A.4]{yin2026emergent}, reporting overall success rate, first-try success rate, and throughput, with the differences noted below.

\paragraph{Initialization.} We sample initial object configurations by uniformly randomizing object positions over the NIST board. The board is fixed to the table with command strips. In simulation the receptive objects are static, and the policy learns to reorient the insertive object relative to a fixed reference. If the board is not secured, the receptive objects can shift during contact, which causes failures.

\paragraph{Trials.} We run the policy continuously for approximately 83 minutes on the nut, 42 minutes on the rod, and 15 minutes on the gear, collecting 48, 49, and 50 trajectories, respectively.

\paragraph{Metrics.} A trial is a first-try success if the policy completes the insertion on its first attempt. A failed insertion followed by another attempt (e.g., re-poking) counts as a retry. Unlike OmniReset, which counts any re-grasp as a retry, we do not penalize re-grasps that occur before the first insertion attempt. Throughput is the number of successful trajectories divided by the total evaluation time, which includes reset time.

\end{document}